\documentclass[11pt,letterpaper]{article}

\usepackage{tabularx}
\usepackage{caption}
\usepackage{floatrow}%
\usepackage[outercaption]{sidecap}

\usepackage{enumitem} 
\usepackage{nicefrac} 

\usepackage{emnlp2016}
\usepackage{times}
\usepackage{latexsym}
\usepackage{graphicx}
\usepackage{subfigure} 
\usepackage{color}
\usepackage{amsmath}

\emnlpfinalcopy

\definecolor{atomictangerine}{rgb}{0.8, 0.2, 0.1}
\definecolor{turq}{rgb}{0.0, 0.5, 0.5}
\definecolor{bright}{rgb}{0.8, 0.1, 0}

\newcommand\jbdone[1]{{}}
\newcommand\vkdone[1]{{}}
\newcommand\gcdone[1]{{}}

\newcommand\ignore[1]{{}}
\newcommand\sat{{SA\&T}}

\title{Learning to generalize to new compositions in image understanding}

\author{Yuval Atzmon$^1$ \And Jonathan Berant$^{3}$ \And
       Vahid Kezami$^3$ \AND Amir Globerson$^{2,3}$ \and Gal Chechik$^{1,3}$
       \\\\\\
       $^1$Gonda Brain Research Center, Bar Ilan University, Israel\\
       yuval.atzmon@biu.ac.il\\ 
       $^2$Tel Aviv University, Israel\\
       $^3$Google Research, Mountain View CA, USA\\
       }

\date{}
\begin{document}
\maketitle

\begin{abstract}
  Recurrent neural networks have recently been used for learning to describe images using natural language. However, it has been observed that these models generalize poorly to scenes that were not observed during training, possibly depending too strongly on the statistics of the text in the training data. 
  Here we propose to describe images using short structured representations, aiming to capture the crux of a description. These structured representations allow us to tease-out and evaluate separately two types of generalization: standard generalization to new images with similar scenes, and generalization to new combinations of known entities. 
  We compare two learning approaches on the MS-COCO dataset: a state-of-the-art recurrent network based on an LSTM (Show, Attend and Tell), and a simple structured prediction model on top of a deep network. We find that the structured model generalizes to new compositions substantially better than the LSTM, $\sim\!\!\!\!7$ times the accuracy of predicting structured representations. By providing a concrete method to quantify generalization for unseen combinations, we argue that structured representations and compositional splits are a useful benchmark for image captioning, and advocate compositional models that capture linguistic and visual structure.
  
\end{abstract}

\section{Introduction}

\begin{figure}[ht]
  \begin{center}
  \includegraphics[scale=0.48, trim={4.5cm 0 4.5cm 0},clip]{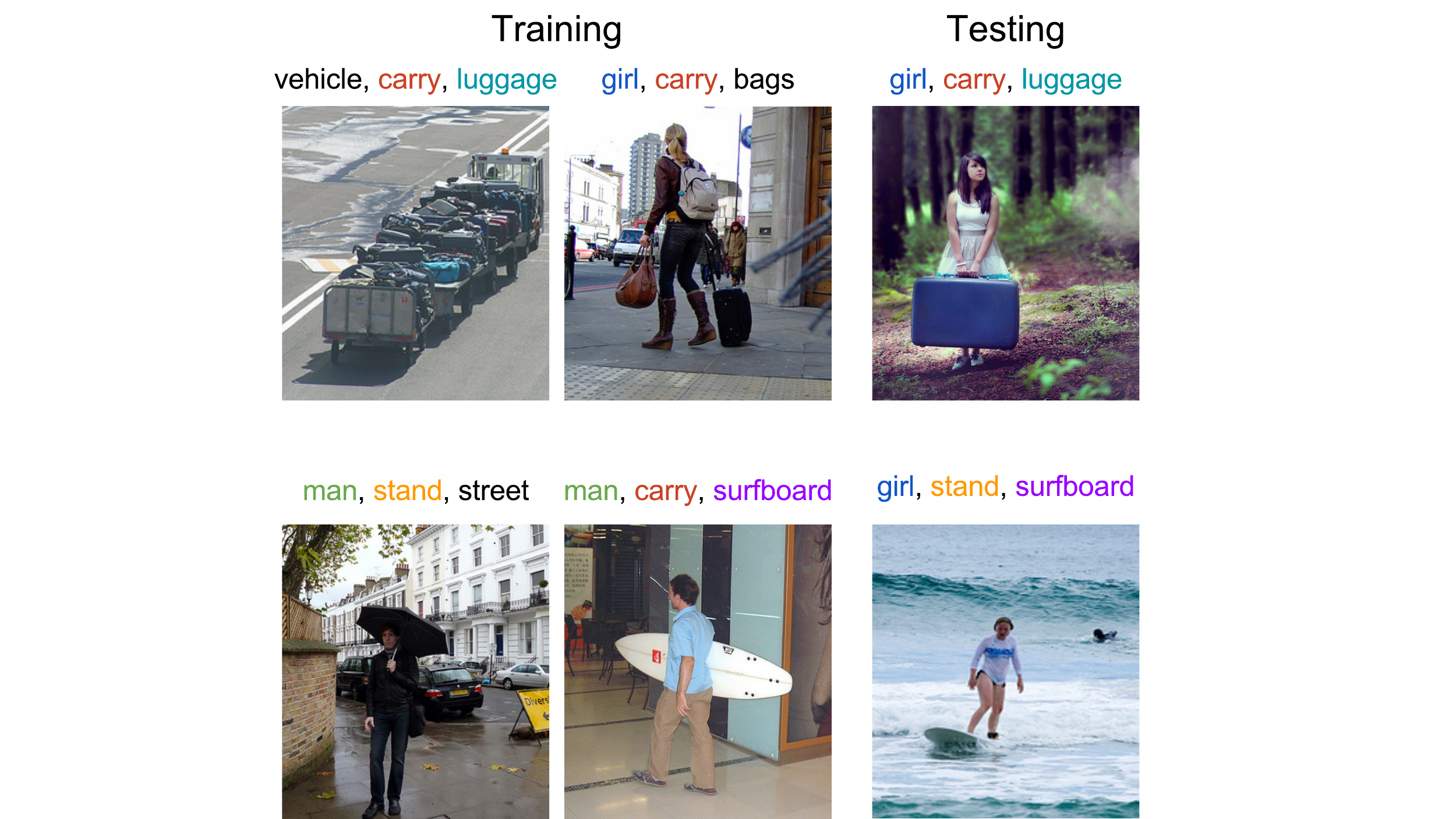}
   \caption{Our motivating task: Learning to generalize to new compositions of entities in images, reflected in their descriptions. Each image is represented with subject-relation-object (SRO) tuple. In a compositional split, testing is performed over novel compositions of entities observed during training, namely, all images matching a given SRO are assigned either to training or testing.}
   \label{compSplit}
   \end{center}
\end{figure}

Training models that describe images with natural language embodies fundamental problems in both language and image understanding. It allows to ground the meaning of language in visual data, and use language compositionality to understand rich visual scenes. 
Recently, deep neural networks have been successfully used for this task \cite{coco2015challenge}. While the results were both inspiring and impressive, it became clear in the aftermath of analyzing the results, that current approaches suffer from two fundamental issues. First, generalization was poor for images describing scenarios not seen at training time.
Second, evaluating descriptions was challenging, because strong language models can generate sensible descriptions that are  missing essential components in the image.
However, a quantitative evaluation of these two problems is still missing. 

In this paper, we propose to address these issues by focusing on structured representations for image descriptions. As a first step, we use simple structured representations consisting of subject-relation-object (SRO) triplets \cite{farhadi2010every}. 
 By reducing full sentences to an SRO representation, we focus on the composition of entities in an image. This has two main advantages. First, it allows to quantify the quality of model predictions directly using the accuracy of SRO predictions. Second, it allows to partition the data such that the model is tested only on new combinations, which are not included in the training set. This allows to evaluate compositional generalization to unseen scenarios, as illustrated in Figure \ref{compSplit}.

We partition the MS-COCO dataset using a compositional split and compare a state-of-the-art recurrent attention model, Show-Attend-and-Tell, \cite{xu2015show} to a structured prediction model built on top of a deep CNN. The recurrent model achieves similar performance on the traditional MS-COCO split. However, we find that it only achieves $\sim 14\%$ the accuracy of the structured model when tested on the new partitioning that requires generalization to new combinations. 

\vspace{-4pt}
\section{Generalizing to novel compositions}
\label{sec:compositionality}
Our key observation is that one should separate two kinds of generalization that are of interest when generating image descriptions. The first, generalizing to new images of the same class, is routinely being evaluated, including in the current data split of the MS-COCO challenge \cite{lin2014microsoft}. The second type, which we focus on, is concerned with generalizing to new scenarios, akin to 
{\em transfer} or {\em zero-shot} learning \cite{fei2006one}, where learning is extended to semantically-similar classes. Importantly, this generalization is the crux of learning in complex scenes, since both language and visual scenes are compositional, resulting in an exponentially large set of possible descriptions. Hence, a key goal of learning to describe images would be to properly quantify generalization to new combinations of known entities and relations.
 
To tease out {\bf{compositional generalization}} from standard within-class generalization, we propose to construct a test set that only contains scenarios that never appeared in the training data. 

In practice, we first map image descriptions to short open-IE style phrases of the form subject-relation-object (termed {\em SRO triplets}). We then partition the examples such that the test and training sets share no common images or SRO triplets (see Figure \ref{compSplit}). This {\bf compositional split} is a natural way to test generalization in short utterances of natural language, since a small training set could be used to train for the large set of possible combinations at test time. 
While some overlap between scenarios in the training and test set can still occur due to synonymy, we hypothesize that this partitioning leads to a much stronger need for generalization.

\vspace{-4pt}
\section{A Structured Prediction Model} \label{sec:model}
To jointly predict an SRO triplet, we train a structured-prediction model on top of a deep convolutional network. First, an image is analyzed to produce candidate bounding boxes \cite{erhan2014scalable} with their labels \cite{Szegedy_2015_CVPR}. Similar to \newcite{xu2015show}, the classifier was trained on a large dataset without fine-tuning on the current data. 

For the structured model on top of the deep network, we used {\em structured SVM} (SSVM) \cite{ssvm2005}, minimizing the hinge loss between the predicted and ground-truth SRO triplets. Specifically, our model learns a score function $f(s,r,o)$ on SRO triplets, decomposed as:
\begin{eqnarray*}
    f(s,r,o) &\!=\!& w_S f_S(s) + w_O f_O(o) + w_R f_R(r) + \\
          & & w_{SR} f_{SR}(s,r) + w_{RO} f_{RO}(r,o),
\end{eqnarray*}
where $w_S, w_O, w_R, w_{SR}, w_{RO}$ are scalar weights learned by the algorithm.
Here, $f_S(s)$ is a score assigned to the subject $s$,  $f_O(o)$ is a score assigned to the object, $f_R(r)$ is a score assigned to the relation, $f_{SR}(s,r)$ is the binary feature over the subject and relation and similarly for $f_{RO}(r,o)$. For details of the model see Appendix A. To get a better understanding of the signals that are useful for the SRO prediction we experimented with multiple variants for the model potentials, for details see section \ref{sec:ssvm_models}

\section{Experiments}
\vspace{-5pt}
\subsection{The Data}
\vspace{-5pt}
We evaluated image captioning on the MS-COCO data \cite{lin2014microsoft}, currently the standard benchmark for evaluating image captioning models (328K images, $\le 5$ textual descriptions per image). 
We parsed MS-COCO descriptions into SRO triplets by first constructing dependency parse trees for each description \cite{andor2016globally}, and then using manually-constructed patterns to extract triplets from each description. Finally, each word was stemmed. Removing descriptions without SROs (due to noun phrases, rare prepositions, or parsing errors), 
yielded 444K unique (image, SRO) pairs \footnote{The templates and SRO triplets are available online at http://chechiklab.biu.ac.il/\textasciitilde yuvval/CompCRF}. 

Analyzing structured phrases and images naturally involves grounding entities to specific image locations. Datasets like Visual-Genome \cite{krishna2016visual} and MS-COCO provide human-marked bounding boxes for many entities. Here, with the goal of being able to generalize to new entities and larger datasets, we instead inferred bounding boxes using a pre-trained deep-network localizer \cite{erhan2014scalable}. We limited nouns to a vocabulary from the 750 most frequent nouns, selecting the $300$ entities that were localizable. and the vocabulary of relations to the top 50 relations, yielding 136K SRO triplets. 

The vocabulary of the visual entity recognition used by the localizer does not fully overlap the the vocabulary of captions. For instance, the term \textit{``cow''} may appear in the captions, while the terms \emph{\{``ox'', ``bull''} and  \emph{ ``calf''\}} may obtain high scores by the localizer.
To match the two vocabularies we followed the procedure of \newcite{zitnick2013learning}, see Appendix B for details.
This mapping was used to select images whose predicted entities matched entities in the captions. 
When an image had several bounding boxes for the same label, we selected the one with the highest score. We also removed duplicate triplets per image, and triplets where the subject and object have the same bounding box. After keeping only images with bounding boxes for both subject and object we were left with 21,213 (image, SRO) pairs with 14,577 unique images .

This dataset was split in two ways: by intersecting with the COCO benchmark split, and in a compositional way as described in Section \ref{sec:compositionality}.

\subsection{Compared Methods}
We compared the following methods and baselines:

\begin{enumerate}
    \item \textbf{SSVM/Conv} Our model described in Sec. \ref{sec:model}.
    
    \item \textbf{Show-Attend-and-Tell (\sat)}. A state-of-the-art RNN attention model for caption generation \cite{xu2015show}. We re-trained the decoder layers to predict SRO triplets with soft-attention. Hyper-parameters were tuned to maximize accuracy on an evaluation set, learning rate in $(0.1, 0.05, 10^{-1}, 10^{-3})$ and weight decay in $(0, 10^{-8}, 10^{-7}, \dots,  10^{-2})$. Importantly, we also controlled for model capacity by tuning the embedding dimensionality $(100, 200, 400, \dots, 1600 \text{ and the default } 512)$ and the LSTM dimensionality $(2^6, 2^7, \dots, 2^{11})$ See Section \ref{sec:results}. The remaining parameters were set as in the implementation provided by \newcite{xu2015show}.
    
    \item \textbf{Stochastic conditional (SC)}. Draw $R$ based on the training distribution, then draw $S$ and $O$ based on the training distribution $p_{train}(S|R)$, $p_{train}(O|R)$. This baseline is designed to capture the gain that can be attributed to bigram statistics. 
    
    \item \textbf{Most frequent triplet (MF)}. Predict an SRO consisting of the most frequent subject, most frequent relation, and most frequent object, based on the training set distribution. By construction, by the way the compositional split is constructed, the most frequent full SRO triplet in the training set can not appear in the test set.
\end{enumerate}

\subsection{Evaluation procedure}
We test all candidate pairs of bounding boxes (BB) for an image. For each BB pair, all candidate SRO triplets are ranked by their scores and compared against the set of ground-truth SRO triplets to compute precision@$k$ for that image. Images may have more than one ground-truth SRO since they are associated with up to $5$ descriptions. For image captioning, BLEU score is a common metric. Here,  SRO-accuracy is equivalent to BLEU-3, and single-term accuracy is equivalent to BLEU-1. We found computing BLEU between a description and its SRO to be too noisy.

Our evaluation metric does not handle semantic smearing, namely, the case where an image can be described in several ways, all semantically adequate, but using different words and hence counted as errors. This issue is often addressed by representing words in continuous semantic spaces. For keeping this paper focused, we leave this outside of current evaluations

We experimented with two cross-validation procedures. First, \textbf{COCO split}, we used the train-test split provided by ms-coco, restricted to the set of images with SROs (COCO split). Second, \textbf{Compositional split}, was applied to unique SRO triplets to create a (80\%/20\%) 5 fold cross validation split. Any object or subject that did not appear in the train set, were moved from the test to the training set with all their triplets (since otherwise they cannot be evaluated). When an object or a subject class appeared only on the test set, then its triplets were moved to the train set. Subject or object appearing less than 5 times were removed from training set. The same (random) set of images was used across all approaches. The fraction of images sometimes deviates from (80\%/20\%) since some triplets have more images than others.

\subsection{Results} \label{sec:results}

\begin{figure}[t]
\begin{center}
  \centerline{\bf{COCO split} \hspace{150pt}}
  \includegraphics[width=7.1cm]{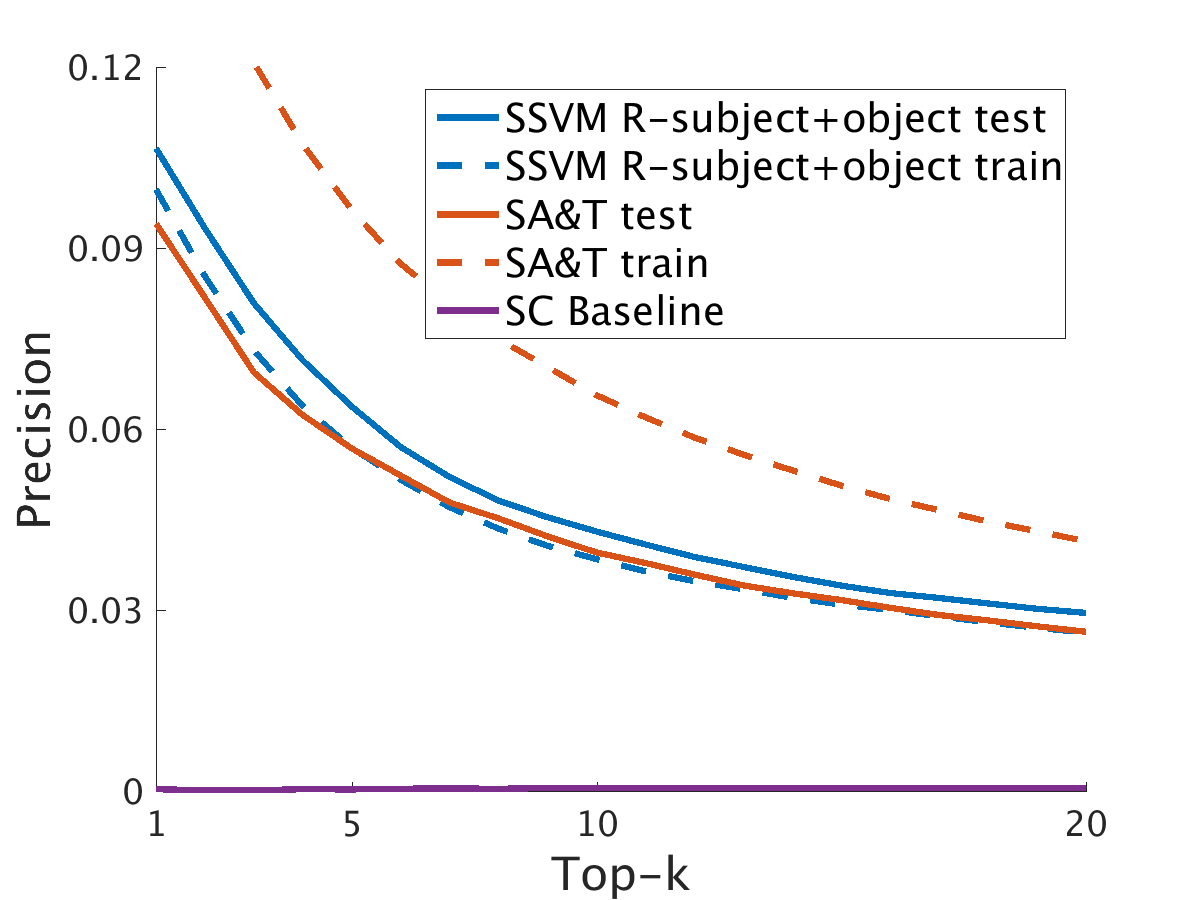}
  \centerline{\bf{Compositional split} \hspace{120pt}}
   \includegraphics[width=7.1cm]{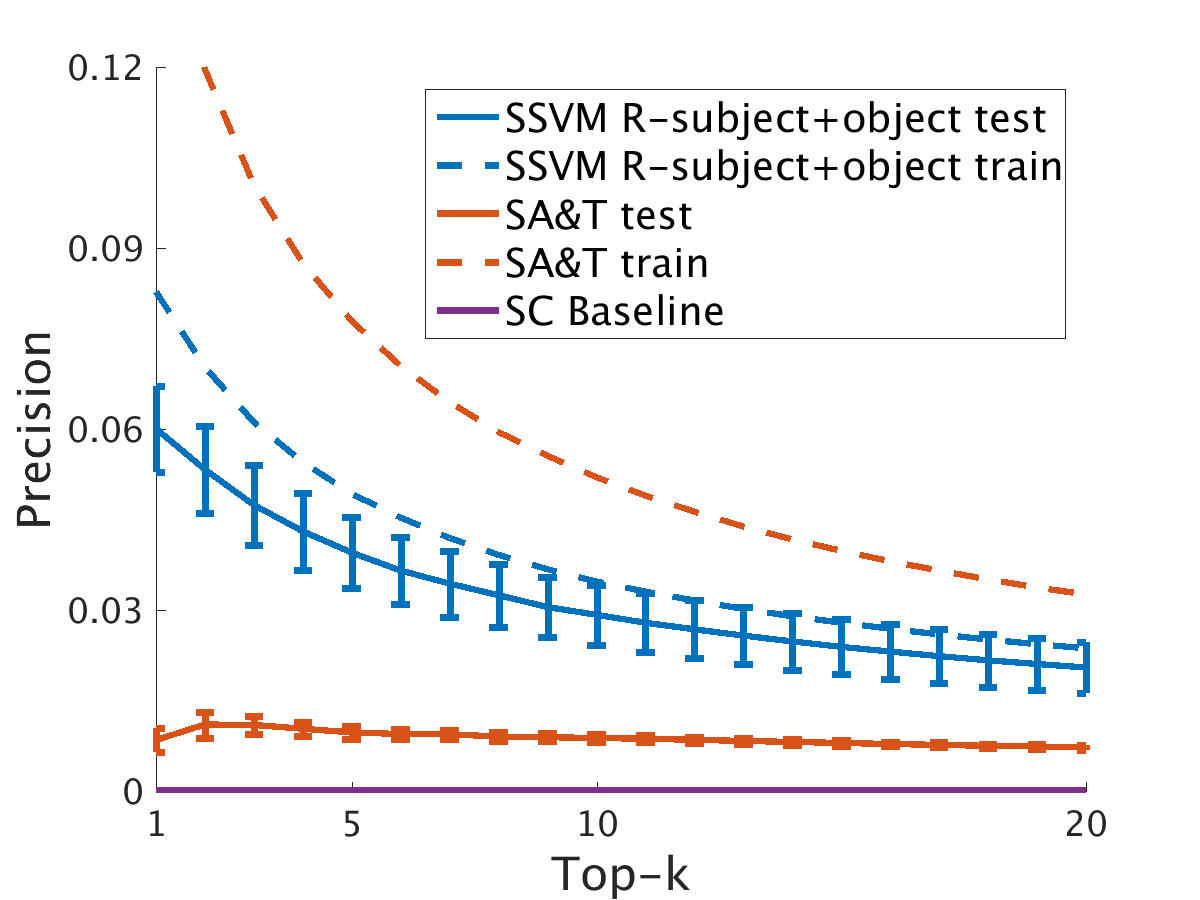}
   \caption{Comparing \sat{} with SSVM/conv. (a) MS-COCO split. (b) Compositional split. \sat{} overfits more strongly than SSVM on the compositional split. Error bars denote the  standard error of the mean (SEM) across five CV folds.}
   \label{fig-lstm-vs-ssvm}
   \end{center}
\end{figure}

\subsubsection{Compositional vs. within-class generalization}
Figure \ref{fig-lstm-vs-ssvm} and Table 1 show average precision@$k$ across images, comparing SSVM to \sat{} for both their test and training performance. In the top panel, both methods are trained and tested on the MS-COCO split. The SSVM/Conv model (blue) wins with precision of $p@1=10.6\%$ and the \sat{} model (green) achieves  $p@1=9.4\%$. Test precision of the baselines was $p@1= 0.028\%$ for SC. The most frequent S, R and O in the dataset were {\em man}, {\em with} and {\em table}, but the triplet (man with table)  did not appear at all in the data, yielding $0\%$ MF accuracy.

The effect is noticeable for the compositional split (bottom panel). Here, the SSVM/conv model transfers well to new combinations (compare training $p@1=8.3\%$ and test $p@1=6\%\pm0.7\%$). Importantly, \sat{} dramatically fails on new combinations, with a large generalization gap as apparent by the difference between precision on the training set ($p@1=15\%$) and the test set only ($p@1=0.85\%\pm0.2\%$). 
Test precision of the baselines was $p@1= 0.014\%$ for SC, and $0\%$ for MF.


\subsubsection{Model complexity}
Generalization gap is often due to over-fitting, which can be controlled by reducing model capacity. We therefore tested \sat{} with different capacities, varying the number of parameters (word dimensionality and LSTM hidden state dimensionality). As expected, training error decreased with the number of parameters. Importantly, test error decreased up to some point and then started rising due to over-fitting. For the MS-COCO split, the \sat{} best test error was better than the SSVM model, but for the compositional split it was significantly worse. In other words, A wide range of LSTM parameters still does not generalize well to the compositional split. Importantly, the number of examples in our experiments is well within the range of dataset sizes that \sat{} was originally used in ({\small Flickr8k, Flickr30k, COCO}). At the same time the SSVM model is limited to bigram potentials, and as such unable to memorize SRO triplets, which the LSTM model may do. We conclude that merely reducing the capacity of the \sat{} model was not sufficiently effective to control overfitting for the compositional case.

\begin{table*}[ht!]
\vskip 0.15in
\begin{small}
\begin{tabular}{lllll}
\hline 
           & \vline \,\,\,\, Compositional & \!\!\!\!\!\!\! Split  & \vline \,\,\,\, COCO  Split & \\ 
\hline
Method     & \vline \, Prec@1     & Prec@5 &  \vline \, Prec@1  \,\,\,\,\,\,\,\,\,\,\,\,\,  & Prec@5  \\
\hline
SSVM R-subject+object & \vline \, \textbf{6.0} &    4.0 & \vline \, \textbf{10.6} & \textbf{6.4} \\
SSVM R-object & \vline \, 5.7 &    \textbf{4.2} & \vline \, 8.3 & 5.6 \\
SSVM R-subject & \vline \, 5.7 &    3.1 & \vline \, 8.8 & 4.8 \\
SSVM no relation features    & \vline \, 4.8 &    3.5 & \vline \, 4.3 & 2.2 \\
SSVM R-spatial+object   & \vline \, 4.3 &    3.2 & \vline \, 5.6 & 3.2 \\
SSVM R-spatial     & \vline \, 4.0 &    2.1 & \vline \, 3.9 & 2.0 \\
\hline
Show Attend \& Tell (SA\&T)     & \vline \, 0.85 &    1.0 & \vline \, 9.4 & 5.7 \\
Stochastic Conditional (SC) & \vline \, 0.014 &  0.018 & \vline \, 0.028 & 0.025 \\
Most Frequent (MF)         & \vline \, 0 &    0 & \vline \, 0 & 0 \\
\caption{{\bf{Ablation experiments.}} Precision@k results (in \%) of the tested methods for the compositional split and the COCO split.
}
\captionsetup{font=small}

\end{tabular}
\end{small}

\vskip -10pt
\end{table*}

\subsubsection{Comparing SSVM models}
\label{sec:ssvm_models}

To get a better understanding of the signals that  are useful for the SRO prediction, we compared multiple variants of the SSVM model, each using different features as the $R$-node potential inputs, for details on the potentials see Appendix A.

\begin{enumerate}
    \item \textbf{SSVM R-subject+object}: 
    The R node potential takes the object (O) category and subject (S) category, each is represented as a sparse "one-hot" vector.
    \item \textbf{SSVM R-object}: 
    The R node potential takes only the object (O) category, represented as a sparse "one-hot" vector.
    \item \textbf{SSVM R-subject}:
    Same for the subject (S), again represented as a sparse "one-hot" vector.
    \item \textbf{SSVM spatial}:
    The R node potential inputs include only spatial features.
    \item \textbf{SSVM R-spatial+object}:
    Inputs include both the spatial features and the object category represented as a one-hot vector. 

    \item \textbf{SSVM no relation features}:
    The R node potential takes no input features, and is only based on the labels frequencies of R in the training set.
\end{enumerate}

Table 1 compares the performance of these models. The best performance is achieved when only taking the predicted labels of the object and subject as  input features for the R node potential. These results suggest that the information in the spatial features is small compared to information in the labels predicted from the pixels.

\subsubsection{Manual evaluation} 
Since images can be described in myriad ways, we manually sampled 100 random predictions of the SSVM model to assess the true model accuracy. For every SRO prediction we answered two questions: (a) Does this SRO exist in the image (b) Is this a reasonable SRO description for the image. In 32\% of the cases, SSVM produced an SRO  that exists in the image, and 23\% of the cases it  was a reasonable description of the image.

\section{Related Work}
Automatic description of images was developed by several groups
\cite{xu2015show,karpathy2015deep,mao2014explain,kiros2014unifying,donahue2015long,vinyals2015show,venugopalan2014translating,chen2014learning,fang2015captions},
and was also applied to parts of images  \cite{johnson2015densecap,krishna2016visual}.

Compositional aspects of language and images have been recently explored by \cite{andreas2015deep}, who approached a visual QA task by breaking questions into substructures, and re-using modular networks. \cite{johnson2015image} combined subjects, objects and relationships in a graph structure for image retrieval. \cite{kulkarni_baby_2011} learned spatial relations for generating descriptions based on a template. \cite{zitnick2013learning} modelled synthetic scenes generated using CRF. The dataset of \cite{yatskarsituation} has combinations of entities modelled with CRF. \cite{farhadi2010every} developed ways to match sentences and images, through a space of meaning parametrized by subject-verb-object triplets which our structured model is closely related to.  Very recently, \cite{lu2016visual} trained a model that leverages language priors from semantic embeddings to predict subject-relation-object tuples. The performance of their model on the unseen-compositions subset in their test set, exhibits a very large generalization gap. Finally, generalization to new objects has often been achieved by ``smearing'' to semantically-related entities \cite{frome2013devise,andreas2015deep,xian2016latent}, but this is outside the scope of this paper. 

\section{Summary}
This paper has two main contributions. First, we  highlight the role of generalization to new combinations of known objects in vision-to-language problems, and propose an experimental framework to measure such compositional generalization. Second, we find that existing state-of-the-art image captioning models generalize poorly to new combinations compared to a structured-prediction model.
In future work, we plan to extend our approach to full captions and handle deeper semantic structures, including modifiers, adjectives and more.

\section*{Appendix A: A structured-SVM model}
Our model learns a score function $f(s,r,o)$ on SRO triplets, decomposed as:
\begin{eqnarray*}
&& w_S f_S(s) + w_O f_O(o) + w_R f_R(r) + \\
&& w_{SR} f_{SR}(s,r) + w_{RO} f_{RO}(r,o),
\end{eqnarray*}
where $w_S, w_O, w_R, w_{SR}, w_{RO}$ are scalar weights learned by the algorithm.
\newline

\noindent{\textbf{Subject node potential $f_S(s)$}}. We learned a sparse linear transformation matrix from the localizer vocabulary to the caption entities vocabulary, bases on empirical joint probability on training data. For example, $f_S({\emph{"cow"}})$ was learned to be a weighted combination of the likelihood scores that the localizer gives to the classes \emph{\{``ox'', ``bull'', ``calf''\}}.
    
\noindent{\textbf{Object node potential $f_O(o)$}}. The $f_O(o)$ potential is defined similarly to $f_S(s)$.
    
\noindent{\textbf{The relation node potential $f_R(r)$}}. The relation node was trained in a separate stage using the same train-test folds, as follows. A multiclass SVM is trained to predict $R$ from features of the subject and object bounding boxes. At inference time, $f_R(r)$ is set as the score that the SVM assigns to relation $r$ in the given image. 
    For input features on some experiments (Section \ref{sec:ssvm_models}), we used the subject or object one-hot-vector or both. Each one-hot-vector is 300 features.
    For spatial features we use the following:
    \begin{itemize}[noitemsep] 
        \item The position, dimension and log dimension of the two boxes $(height, width, x, y)$.
    \item The distance and log distance of a vector connecting the center of the subject box with that of the object.
    \item The angle of a vector connecting the center of the subject box with the object box, represented as a x,y pair normalized to length 1.
    \item Aspect ratio combinations of box dimensions, including $h_S/w_S, h_S/h_O$ and similar ratios.
    \item The square root of box areas, and the ratio and log-ratio of square root box areas.
    \item The area of intersection and the intersection over union.
    \item The square root relative overlap of the subject-object areas $(\text{intersect(SO)}/\text{area( O)})^{\frac{1}{2}}$. Similarly for object-subject.
    \item Binary conditions, including 
    \begin{itemize}[noitemsep] 
    \item $\text{Relative overlap (SO)} < 0.25$ 
    \item $\text{Relative overlap (OS)} < 0.25$ 
    \item $\text{Relative overlap (OS)} > 0.85$ 
    \item $x_S < x_O$ 
    \item $y_S < y_O$ 
    \item $(y_S < y_O)\text{ and }(x_S < x_O)$ 
    \item $(y_S < y_O)\text{ and }\text{not}(x_S < x_O)$ 
    \end{itemize}
    \end{itemize}
    The spatial features were then normalized to zero mean and unit variance.  

\noindent{\textbf{The pairwise feature $f_{SR}(s,r)$.}} This potential was set as the bigram probability of the combination $(s,r)$, as estimated from the training data, and similarly for $f_{RO}(r,o)$.

\section*{Appendix B: matching visual entities to caption terms}

When creating the dataset, we selected those images where the visual entities can be mapped to terms in the captions. Since the vocabulary of the visual entity recognition (used by the localizer) differs from the vocabulary of captions, we  estimated a mapping from the locaizer vocabulary to the caption terms following the procedure of \newcite{zitnick2013learning}. 

Specifically, (1) We computed PMI between the labels predicted by the localizer for the bounding boxes (BBLs) and the nouns in the SRO. (2) We considered the top-5 matches for each S/O vocabulary word, and manually pruned outliers (for instance, the term {\em bed} had high MI with {\em cat} detections). (3) We removed a data sample if the S/O caption terms did not match any of the BBLs. This PMI step results in having 300 entities.

This transformation was only used for selecting the subset of the data that contains the set of entities in the $S/O$ vocabulary.

\bibliography{emnlp2016}
\bibliographystyle{emnlp2016}

\end{document}